\def\year{2021}\relax
\documentclass[letterpaper]{article} 
\usepackage{aaai21}  
\usepackage{times}  
\usepackage{helvet} 
\usepackage{courier}  
\usepackage[hyphens]{url}  
\usepackage{graphicx} 
\usepackage[ruled]{algorithm2e}
\usepackage[dvipsnames]{xcolor}
\usepackage{caption} 
\title{Normalized Label Distribution: Towards Learning Calibrated, Adaptable and Efficient Activation Maps}
\author{
    Utkarsh Uppal $^{\ast}$\textsuperscript{\rm 1, 2}, Bharat Giddwani $^{\ast}$\textsuperscript{\rm 3}\\
    
}
\affiliations{
    
    Punjab Engineering College, Chandigarh, India, \textsuperscript{\rm 1}
    VisioLab, \textsuperscript{\rm 2}
    NVIDIA \textsuperscript{\rm 3}

    \{utkarsh@visiolab.io, bgiddwani@nvidia.com\}

}

\begin{document}
\maketitle
\begin{abstract}
The vulnerability of models to data aberrations and adversarial attacks influences their ability to demarcate distinct class boundaries efficiently. The network's confidence and uncertainty play a pivotal role in weight adjustments and the extent of acknowledging such attacks. In this paper, we address the trade-off between the accuracy and calibration potential of a classification network. We study the significance of ground-truth distribution changes on the performance and generalizability of various state-of-the-art networks and compare the proposed method's response to unanticipated attacks. Furthermore, we demonstrate the role of label-smoothing regularization and normalization in yielding better generalizability and calibrated probability distribution by proposing normalized soft labels to enhance the calibration of feature maps. Subsequently, we substantiate our inference by translating conventional convolutions to padding based partial convolution to establish the tangible impact of corrections in reinforcing the performance and convergence rate. We graphically elucidate the implication of such variations with the critical purpose of corroborating the reliability and reproducibility for multiple datasets.

\end{abstract}

\section{Introduction}
Deep Neural Networks (DNN) are being utilized for wide domain of applications including image classification \cite{szegedy2016rethinking}, speech recognition \cite{chorowski2016towards}, tackling climate change \cite{rolnick2019tackling}, and medical image segmentation \cite{giddwani2020csta,islam2019brain}. Despite achieving state-of-the-art performance, DNN's are making progress to be susceptible to data aberrations. The model's predicted probabilistic distributions are largely overconfident, poorly calibrated, and lacking generalization. Certain modifications like Temperature Scaling \cite{laves2019well} and Label Smoothing (LS) \cite{muller2019does} caters probability prediction with better uncertainty, preventing over-confident training, and improving the generalization capacity. 

We discuss disparate classification approaches to address two contrasting issues in real-time processing - poor calibration and lower feature map distribution; considering them for the efficient pipeline, spontaneous video-image rendering, secure model adoption, and decision making based on class prediction. We integrate label smoothing with trivial cross-entropy loss to facilitate a well-separated class distribution and intensify the model's generalization capacity. Furthermore, we incorporate a normalized \cite{ma2020normalized}  soft targets based cross-entropy loss capable of incurring a better uncertainty without jeopardizing the classification model's performance. 

The flexibility and security of DNN's are critical, especially during translation, deployment, and adoption of these architectures to facilitate real-world problem-solving. Henceforth, we conduct the experiments on our skewed stubble burning dataset and on novel mathematical algorithms. Furthermore, to draw out realistic comparisons with the trivial deep learning techniques, we analyze the class separation boundaries and uncertainty on the CIFAR10 dataset. Liu et. al \cite{liu2018partial} proposed a padding approach to reweigh the border pixels of an image, treating the padded areas as eventual holes in the trivial convolution approach. We incorporate partial padding technique in state-of-the-art (SOTA) classification networks to cater to reweighed pixel maps-based convolution and emphasize the flexibility of the label modification techniques. Consequently, to address the geometric repercussions of label distribution alterations, we study the tSNE projections of the penultimate layer distribution for the disparate cost functions.

Our overall contributions can be summarized as follows:
\begin{quote}
\begin{itemize}
\item We corroborate the trade-off and the impact of ground-truth distribution changes to uncertainty, confidence, and generalizability of feature maps prediction for a classification task.    
\item We integrate normalization to label smoothing cross-entropy loss, allowing cost function to impede network’s over-confidence, refine calibration, and enhance the model’s performance and uncertainty capacity.
\item While bearing adversarial attacks or unforeseen hyperparameters in the form of real-time skewed datasets or novel mathematical functions, our proposed approach validate flexible and reproducible performance and attribute grasping through better class separation boundaries in tSNE embeddings.
\end{itemize}
\end{quote}

\begin{figure*}
  \centering
  \includegraphics[width=0.8\linewidth]{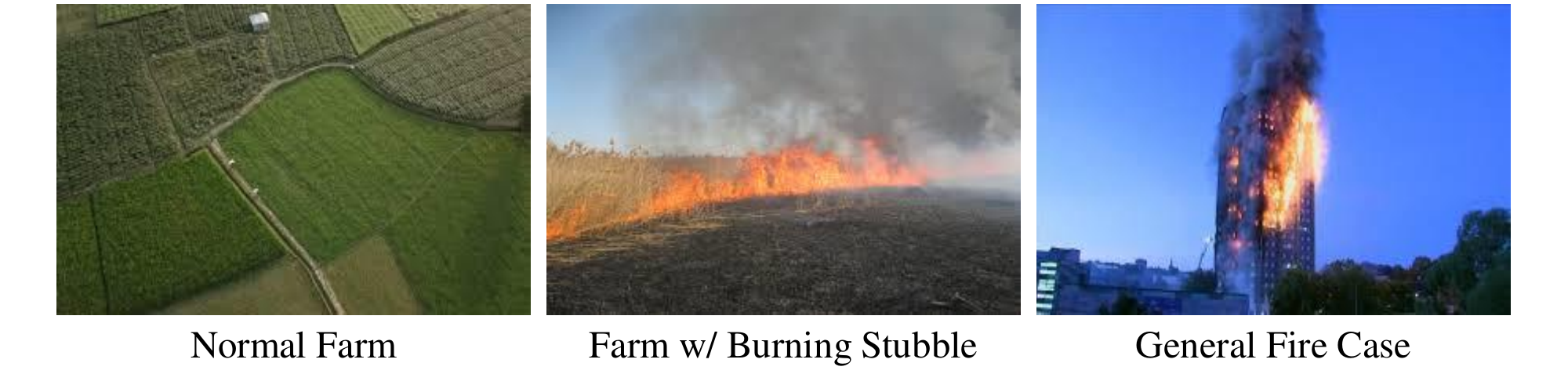}
  \caption{Illustration of sample figures from the skewed dataset. The images represent three distinct classes - regular farm, farm w/ burning stalks, and general fire instance.}
  \label{fig:data}  
\end{figure*}
\begin{table*}
 \caption{Quantitative evaluation of accuracy, precision, recall, and f1-score for SOTA classification models optimized with CE, LSCE, and normalized (norm.) LSCE on our validation dataset. Acc., Prec., Rec., F1. represents Accuracy, Precision and F1-Score respectively.}
  \label{LS_Normal_Ours}%
  \centering
\begin{tabular}{ccccclcccclcccc}
\hline
Model        & \multicolumn{4}{c}{CE (normal targets \%)} &  & \multicolumn{4}{c}{LSCE (soft targets \%)} &  & \multicolumn{4}{c}{norm. LSCE (norm. soft targets \%)} \\ \cline{2-5} \cline{7-10} \cline{12-15} 
             & Acc.    & Prec.  & Rec.   & F1     &  & Acc.   & Prec.  & Rec.   & F1    &  & Acc.   & Prec.  & Rec.   & F1     \\ \hline
ResNet18\cite{he2016deep}      & 78.12   & 90.31  & 78.38  & 83.5   &  & 80.35  & 91.78  & 80.05  & 84.85 &  & 80.8   & \textbf{93.19}  & 83.55  & 86.97   \\ 
ResNet34\cite{he2016deep}      & 80.8    & 92.27  & 78.27  & 83.93  &  & 79.02  & 90.08  & 80.11  & 84.61 &  & 81.25	& 92.26	 & 84.94  & 87.49    \\
ResNet50\cite{he2016deep}      & 74.55   & 90.28  & 69.38  & 76.04  &  & 77.67  & 90.39  & 77.00  & 82.48 &  & 79.91	& 93.14	 & 82.89  & 86.5  \\
MobileNetv2\cite{howard2017mobilenets}  & 81.69   & 92.83  & 79.99  & 85.25  &  & 82.59  & 83.17  & 81.72  & 86.39 &  & \textbf{85.71}	& 93.13	 & \textbf{88.27}  & \textbf{90.28} \\
EfficientNet\cite{tan2019efficientnet} & 75.00   & 89.51  & 59.17  & 60.80  &  & 78.12  & 92.26  & 78.38  & 83.69 &  & 79.02	& 93.07	 & 81.16  & 85.39       \\ \hline
\end{tabular}
\end{table*}
\section{Attacks and Methods}
In this section, we detail the key attributes of the exploratory normalized smooth targets. We also delineate the response of our approach to changes in datasets distribution and network layers to reach SOTA performance while experiencing the real-world environment and set of possibilities.

\subsection{Classification and Label Distribution}
Cross-Entropy (CE) loss function magnifies the probabilistic estimation per class for a multi-class distribution prediction. Eqn \ref{eq-a} defines the CE loss where $\hat{y_c}$ represents the estimated probability and ${y_c}$ corresponds to the one-hot target vector.       

  \begin{equation}
    \label{eq-a}
      \mathcal{L}_\mathrm{CE}= -\sum_{c=1}^N {y_c}\mathit{log(g(\hat{y_c}))}
  \end{equation}
  
CE loss fused with SOTA models has already over-achieved in terms of prediction precision. However, the observed distribution of the data points lacks generalizability and calibration. LS \cite{muller2019does} generates soft targets by re-weighting targets with a uniform distribution. Eqn \ref{eq-b} delineates the Label Smoothing Cross-Entropy (LSCE) loss function, where $y_c^{LS}$ represents the soft targets generated by altering the ground-truth distribution by a uniform distribution sequence of $\frac{\epsilon}{N}$; N representing the total classes and $\epsilon (here, 0.1)$ labeled as the smoothing factor. LSCE engenders better-calibrated prediction and ensues a network with better generalization and uncertainty for distribution demarcation. 
where, $y_c^{LS} = (1-\epsilon) y_c + \frac{\epsilon}{N}$ and $ g(.)$ is the softmax operation.
\begin{equation}
    \label{eq-b}
    \mathcal{L}_\mathrm{LSCE}= -\sum_{c=1}^N  {y_c^{LS}}\mathit{log(g(\hat{y_c}))}
  \end{equation}
Normalization provides an even distribution of the data points to alleviate overfitting and biased behavioral patterns. Eqn. \ref{eq-c} represents the normalized form \cite{ma2020normalized} of LSCE, where the denominator corresponds to the entire range of one-hot target vector. Consequently, the resulting norm. LSCE (normalized Label Smoothing Cross-Entropy) loss is a scaled variant of the calibrated ground-truth vector. 
  \begin{equation}
    \label{eq-c}
    \mathcal{L}_\mathrm{normLSCE}= \frac{-\sum_{c=1}^N  {y_c^{LS}}\mathit{log(g(\hat{y_c}))}} {{-\sum_{j=1}^N \sum_{c=1}^N  {y_{cj}^{LS}}\mathit{log(g(\hat{y_c}))}}} 
  \end{equation}
  
\begin{table*}
 \caption{Quantitative mean evaluation metrics for SOTA classification models optimized with CE, LSCE, and normalized (norm.) LSCE on our validation dataset for partial convolution (PC) based approach. Acc., Prec., Rec., F1. represents Accuracy, Precision and F1-Score respectively.}
  \label{LS_Normal_PC}
  \centering
\begin{tabular}{ccccclcccclcccc}
\hline
Model        & \multicolumn{4}{c}{Partial Conv. (\%)} &  & \multicolumn{4}{c}{Partial Conv. w/ soft targets (\%)} &  & \multicolumn{4}{c}{Partial Conv. w/ norm. soft targets (\%)} \\ \cline{2-5} \cline{7-10} \cline{12-15} 
             & Acc.    & Prec.  & Rec.   & F1     &  & Acc.   & Prec.  & Rec.   & F1    &  & Acc.   & Prec.  & Rec.   & F1     \\ \hline
ResNet18\cite{he2016deep}     & 79.46    & \textbf{92.95}  & 74.11  & 81.46  &  & 78.57  & 91.5	& 75.55	& 81.88 &  & 79.91	&91.44	&79.72	&84.57       \\      
ResNet34\cite{he2016deep}      & 80.36    &92.09   & 76.88  & 82.9  &  & 78.57	&92.06	&79.77	&84.7   &  & 79.46	&92.77	&80.44	&85.31      \\
ResNet50\cite{he2016deep}      & 79.02   & 92.21  & 77.99  & 83.67  &  &79.46	&91.44	&79.38	&84.34   &  &\textbf{81.25}	&92.81	&\textbf{80.77}	&\textbf{85.67}     \\ \hline
\end{tabular}
\end{table*}

\subsection{Dataset Changes}
\begin{figure}
  \centering
  \includegraphics[width=1\linewidth]{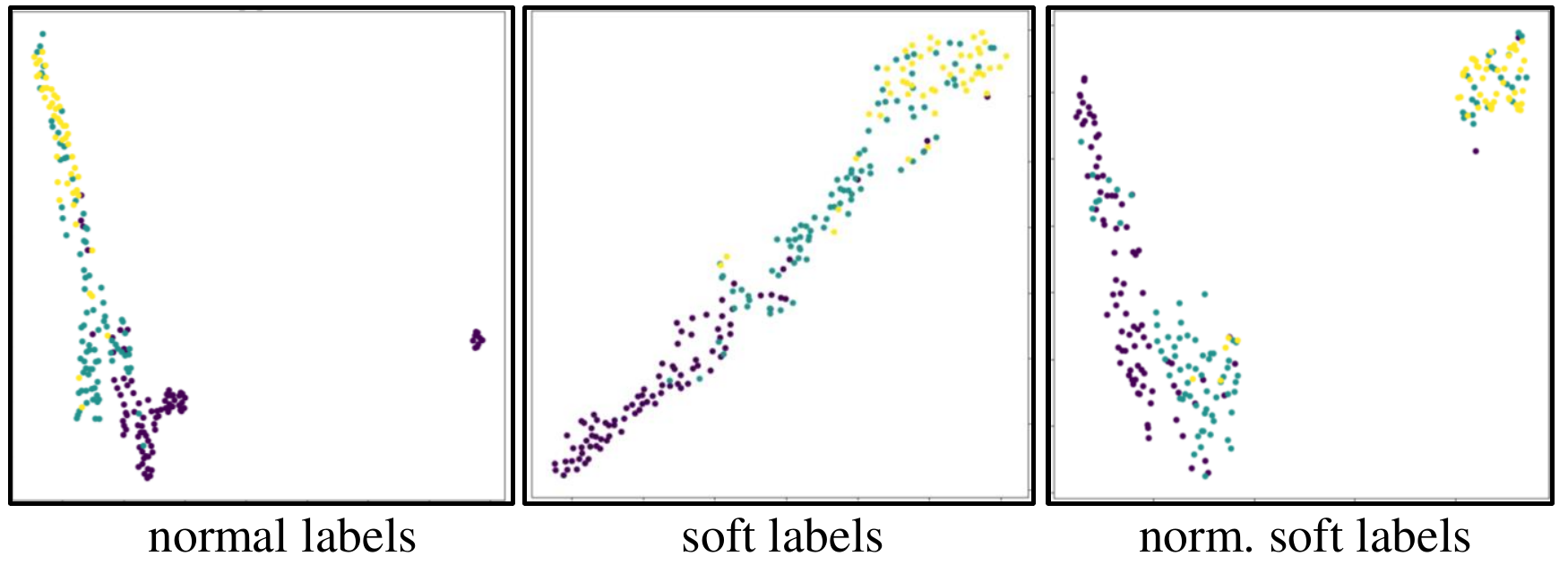}
  \caption{tSNE penultimate layer projections demonstrating class clustering of ResNet18 architecture trained on our dataset for CE, LSCE, norm. LSCE losses.}
  \label{fig:ours_tsne}  
\end{figure}

\begin{figure}
  \centering
  \includegraphics[width=1.0\linewidth]{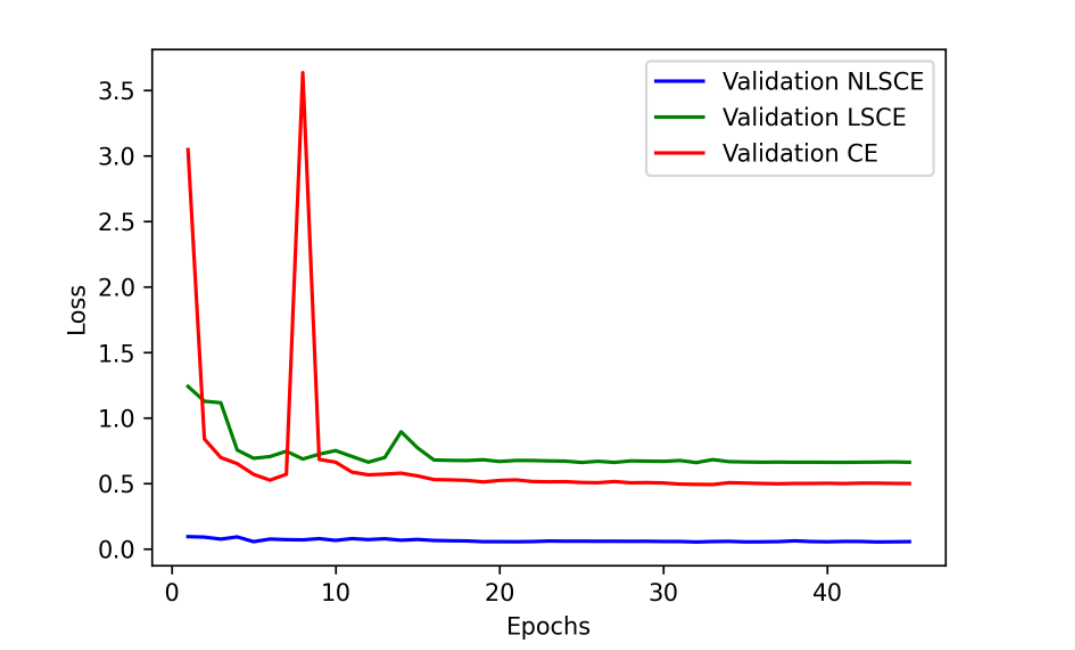}
  \caption{Loss variation plot of validation split across each epoch comparing CE, LSCE, and norm LSCE (NLSCE) for the ResNet18 architecture on our dataset.}
  \label{fig:loss_curve}  
\end{figure}

Specific data outliers can drastically impact the performance of classification models. \cite{szegedy2016rethinking} explains the significance of model uncertainty and the role of smooth target vectors in rendering and amelioration of predicted class labels. In fact, the datasets in practical applications are not ideal and might be skewed, which engenders a confidence bias and irregular predictions. We prepare a benchmark skewed three-class stubble dataset to validate the competence of norm. LSCE  for efficient deployment and performance. The dataset contains 858 images divided into three classes - normal farm, farm w/burning stubble, and generic fire cases. The images have a consistent size of 224x224 and a consistent pixel range [0-255]. Fig. \ref{fig:data} exhibits a sample of images from each of the three classes of our initial dataset.      
We conduct our experiments in Pytorch framework for training and evaluation of several SOTA classification models namely ResNets-18, 34, 50 \cite{he2016deep}, MobileNet \cite{howard2017mobilenets}, and EfficientNet \cite{tan2019efficientnet} to express definite correlation. Table \ref{LS_Normal_Ours} specifies the comprehensive evaluation metrics for models trained with CE, LSCE, and norm LSCE cost function. Fig. \ref{fig:ours_tsne} depicts the tSNE projections of the penultimate layer distribution for disparate cost functions and the better clustering for the soft labels in comparison to the normal labels in case of CE loss. We also compare validation loss-shift curve in Fig. \ref{fig:loss_curve}.


\subsection{Functional Attacks and Alterations}
Partial Convolution (PC) based padding  \cite{liu2018partial} introduces a re-weighted convolution segregating the input pixels as non-hole and zero padding as holes. PC refines the image boundaries based on surrounding pixels, thus returning a better convergence rate while training. We replace the trivial mathematical convolution operation with this padding based convolution in discrete classification networks. Table \ref{LS_Normal_PC} presents similar comparisons but for partial convolution.   

\begin{figure}
  \centering
  \includegraphics[width=1\linewidth]{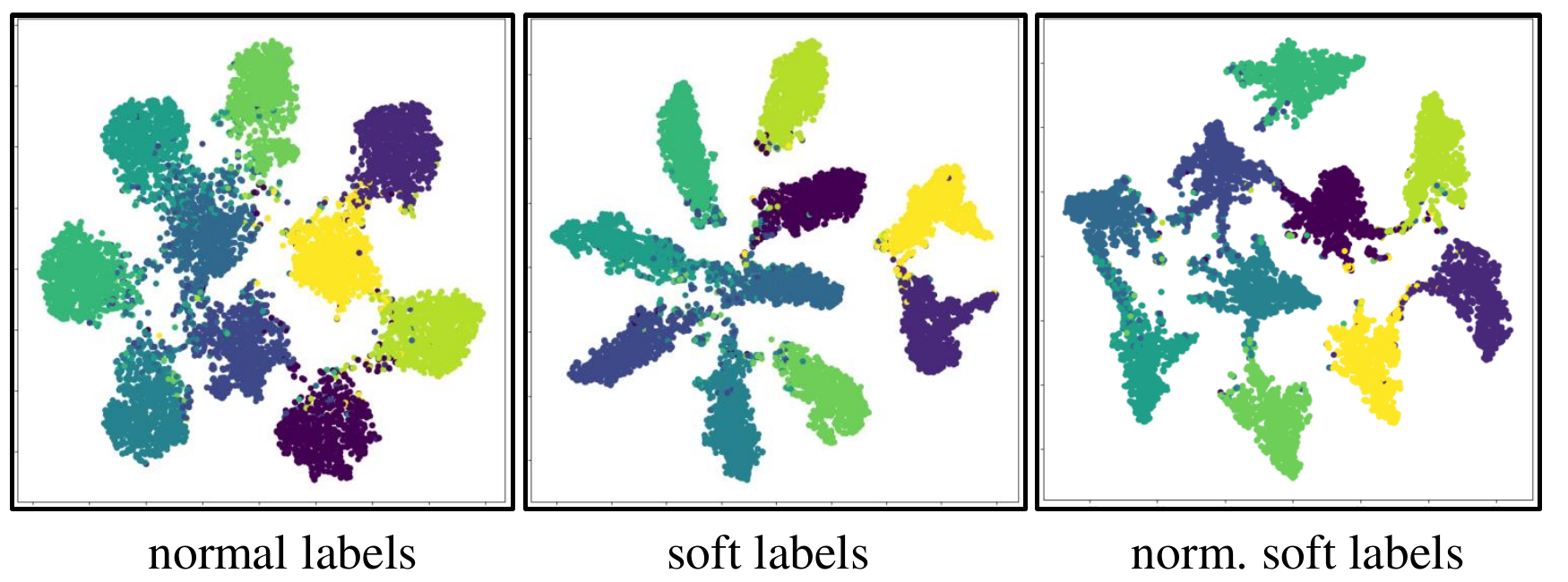}
  \caption{tSNE penultimate layer projections demonstrating class clustering of ResNet18 architecture trained on CIFAR10 dataset for CE, LSCE, norm. LSCE losses.}
  \label{fig:cifar_tsne}  
\end{figure}

To compare the performance and calibration ability with SOTA benchmark accuracy, we extend the dataset and functional attacks experimentation to CIFAR10 dataset \cite{krizhevsky2009learning}. Although the CIFAR10 dataset is well-balanced and equally distributed, the proposed normalization depicts better accuracy and, more importantly, a better confidence distribution and generalization. In table \ref{LS_Normal_CIFAR}, we establish the accuracy correlation for CE, LSCE, norm. LSCE, and norm. LSCE with PC for the CIFAR10 validation set. Fig. \ref{fig:cifar_tsne} depicts the comparisons between the tSNE embeddings for CIFAR10 dataset. The well-separated class boundaries help in the deduction of a calibrated network with better uncertainty.

\begin{table}
\caption{Mean accuracy estimation for various ResNet architectures with CE, Label Smoothing (LS), normLS, normLS (w/ PC) on CIFAR10 validation dataset.}
  \label{LS_Normal_CIFAR}
  \centering
\begin{tabular}{ccccc}
\hline
    & CE    & LS      & normLS & normLS (w/ PC)  \\ \hline
ResNet18\cite{he2016deep} &93.23    &93.88    &\textbf{94.15} &  94.04             \\
ResNet34\cite{he2016deep} &93.50    &93.75    &93.88  &  \textbf{93.94}        \\
ResNet50\cite{he2016deep} &91.53    &91.46    &93.30  &   \textbf{93.42}       \\ \hline
\end{tabular}
\end{table}

\begin{algorithm}[!b]
\DontPrintSemicolon
\SetKwInOut{Input}{Input}\SetKwInOut{Output}{Output}
\Input{\textit{(x,y)}}
    $t = 0$. \;
    \Repeat{$\pi$ {\normalfont converges.}}{
        $\pi^{(t+1)} \leftarrow $ $\pi(x,y)$ w/ ${y_c}$ eqn. (\ref{eq-a})  or \;
        ${\pi_{soft}}^{(t+1)} \leftarrow $ $\pi(x,y)$ w/ soft ${y_c}$ eqn. (\ref{eq-b}) or \;
        ${\pi_{norm. soft}}^{(t+1)} \leftarrow $ Calculate $\pi(x,y)$ w/ normalized soft distribution ${y_c}$ eqn. (\ref{eq-c})\;
        $t \leftarrow t+1$.  \;
    }
\KwRet{$\pi$}.\;
Partial Conv. $\leftarrow$ Change Conv. 
\caption{Normalized Label Smoothing. }
\label{alg:rwm}
\end{algorithm}

\section{Discussions}
With the ever-increasing need to automate real-life processes through the deployment of intelligent algorithms, security breaches, unethical barriers, human errors are chief attributes that might restrict the domain of such algorithms. Notably, for classification tasks encompassing image and pixel-level classification (segmentation), the degree of uncertainty and ability of the model to mitigate data outliers is hugely vital. Adversarial attacks and biased confidence distribution can engender deviations from true behavioral patterns. Hence, robust, secure, and efficient algorithms to overcome such threats are revolutionary to sustain intelligent devices. Trivial label smoothing is reliable in enhancing the calibration of the network. However, the shift to soft targets might not always increase the performance of the network. Consequently, the trade-off between performance and efficiency is a critical attribute in regulating a particular model's behavior. However, the parallel increase in the performance metrics and the well-embarked clustering of classes for normalized label smoothing factor of the cost function can circumvent this issue. In fact, integrating such a mathematical function is a more economical and compact solution in terms of attributes like memory occupancy and hyperparameters of a network. Subsequently, for future research, we can reproduce the values in a comprehensive ablation study to amalgamate the normalized variations with loss functions like focal loss \cite{lin2017focal}, cosine loss \cite{wang2018cosface} into a single consolidated model.

\section{Conclusion}
In our work, we address the susceptibility of classification models to overconfidence, lack of uncertainty, and poor generalization. Consequently, we present a competent and efficient strategy to mitigate this problem by catering to a normalized soft target based cross-entropy loss. We validate the cost function on distinct fluctuations, primarily a skewed dataset and a novel mathematical convolution variant. We corroborate and compare the class-separation boundaries by plotting tSNE projections and illustrating a tighter cluster formation for norm. LSCE function, depicting a better calibrated predicted distribution. Finally, we train the network on the benchmark CIFAR10 dataset to demonstrate norm. LSCE adaptability in capitulating better generalization, uncertainty, class separation, and performance in the model’s predictions.

\bigskip
\noindent 

\bibliographystyle{unsrt}
\bibliography{main}

\end{document}